\documentclass[11pt,a4paper]{article}
\usepackage[hyperref]{emnlp2020}

\usepackage{graphicx}
\usepackage{eurosym}
\usepackage{tablefootnote}
\usepackage{kantlipsum}
\graphicspath{ {images/} }
\usepackage{hyperref}

\usepackage{times}
\usepackage{latexsym}
\usepackage{booktabs}
\usepackage{csquotes}
\usepackage{array}
\usepackage{listings}
\usepackage{stfloats}
\usepackage{hyperref}
\usepackage[super]{nth}
\usepackage[inline,shortlabels]{enumitem}
\usepackage{multirow}
\usepackage{makecell}
\usepackage{siunitx}
\usepackage{multicol}
\usepackage{svg}
\usepackage{float}
\usepackage{ amssymb }
\usepackage{amsmath,amsfonts,amssymb}

\newcounter{notecounter}
\newcommand{\enotesoff}{\long\gdef\enote##1##2{}}
\newcommand{\enoteson}{\long\gdef\enote##1##2{{
			\stepcounter{notecounter}
			\large\bf
			\hspace{100cm}\arabic{notecounter} $<<<$ ##1: ##2
			$>>>$\hspace{1cm}}}}
\enoteson
\enotesoff

\usepackage{microtype}

\aclfinalcopy 

\title{The LMU Munich System for the WMT 2020 Unsupervised Machine Translation Shared Task}

\author{ 
	Alexandra Chronopoulou,
    Dario Stojanovski,
    Viktor Hangya, 
	Alexander Fraser\\\\
	Center for Information and Language Processing, LMU Munich, Germany \\
	{\tt \{achron, stojanovski, hangyav, fraser\}@cis.lmu.de}
	}
\date{}

\begin{document}
\maketitle
\begin{abstract}
This paper describes the submission of LMU Munich to the WMT 2020 unsupervised shared task, in two language directions, German$\leftrightarrow$Upper Sorbian. Our core unsupervised neural machine translation (\textsc{unmt}) system follows the strategy of \citet{chronopoulou2020reusing}, using a monolingual pretrained language generation model (on German) and fine-tuning it on both German and Upper Sorbian, before initializing a \textsc{unmt} model, which is  trained with online backtranslation. Pseudo-parallel data obtained from an unsupervised statistical machine translation (\textsc{usmt}) system is used to fine-tune the \textsc{unmt} model. We also apply BPE-Dropout to the low-resource (Upper Sorbian) data to obtain a more robust system.
We additionally experiment with residual adapters and find them useful in the Upper Sorbian$\rightarrow$German direction. We explore sampling during backtranslation and curriculum learning to use \textsc{smt} translations in a more principled way. Finally, we ensemble our best-performing systems and reach a BLEU score of 
$32.4$
on German$\rightarrow$Upper Sorbian and 
$35.2$
on Upper Sorbian$\rightarrow$German. 
\end{abstract}

\section{Introduction}
Neural machine translation achieves remarkable results \cite{bahdanau2015, vaswani2017attention} when large parallel training corpora are available. However, such corpora are only available for a limited number of languages. \textsc{unmt} addresses this issue by using monolingual data only \cite{artetxe2017unsupervised,lample2018phrase}. The performance of \textsc{unmt} models is further improved using transfer learning from a pretrained cross-lingual model \cite{lample2019cross,song2019mass}. However, pretraining also demands large monolingual corpora for both languages. Without abundant data, \textsc{unmt} methods are often ineffective \cite{guzman2019flores}. Therefore, effectively translating between a high-resource and a low-resource language,  in terms of monolingual data, which is the target of this year's unsupervised shared task, is challenging. 

We participate in the WMT 2020 unsupervised machine translation shared task. The task includes two directions: German$\rightarrow$Upper Sorbian (\texttt{De}$\rightarrow$\texttt{Hsb}) and Upper Sorbian$\rightarrow$German (\texttt{Hsb}$\rightarrow$\texttt{De}). Our systems are constrained,  using only the provided \texttt{Hsb} monolingual data and \texttt{De} NewsCrawl monolingual data released for WMT.
We pretrain a monolingual encoder-decoder model on a language generation task with the Masked Sequence to Sequence model (\textsc{mass})  \cite{song2019mass} and fine-tune it on both languages of interest, following \citet{chronopoulou2020reusing}. We then train it on \textsc{unmt}, using online backtranslation. 
We use our \textsc{usmt} system to backtranslate monolingual data in both languages. This pseudo-parallel corpus serves to fine-tune our \textsc{unmt} model.  
Iterative offline backtranslation is later leveraged, yielding a performance boost. We use BPE-Dropout \cite{bpedropout} as a data 
augmentation
technique, sampling instead of greedy decoding in online backtranslation, and curriculum learning to best include the \textsc{smt} pseudo-parallel data. We also use residual adapters \cite{Houlsby2019ParameterEfficientTL} to translate to the low-resource language (\texttt{Hsb}).  

\noindent 
\textbf{Results Summary.}  
The ensemble of our best-performing systems yields the best performance in terms of BLEU\footnote{\url{http://matrix.statmt.org/matrix/systems_list/1920}} among the participants of the unsupervised machine translation shared task. 
We release the code and our best models\footnote{\url{https://github.com/alexandra-chron/umt-lmu-wmt2020}} in order to facilitate reproduction of our work and experimentation in this field. We note that we have built upon the \textsc{mass} codebase\footnote{\url{https://github.com/microsoft/MASS}} for our experiments.

\section{Model Description} \label{sec:system}
\begin{figure*}
	\centering
	\includegraphics[width=1\textwidth, page=1]{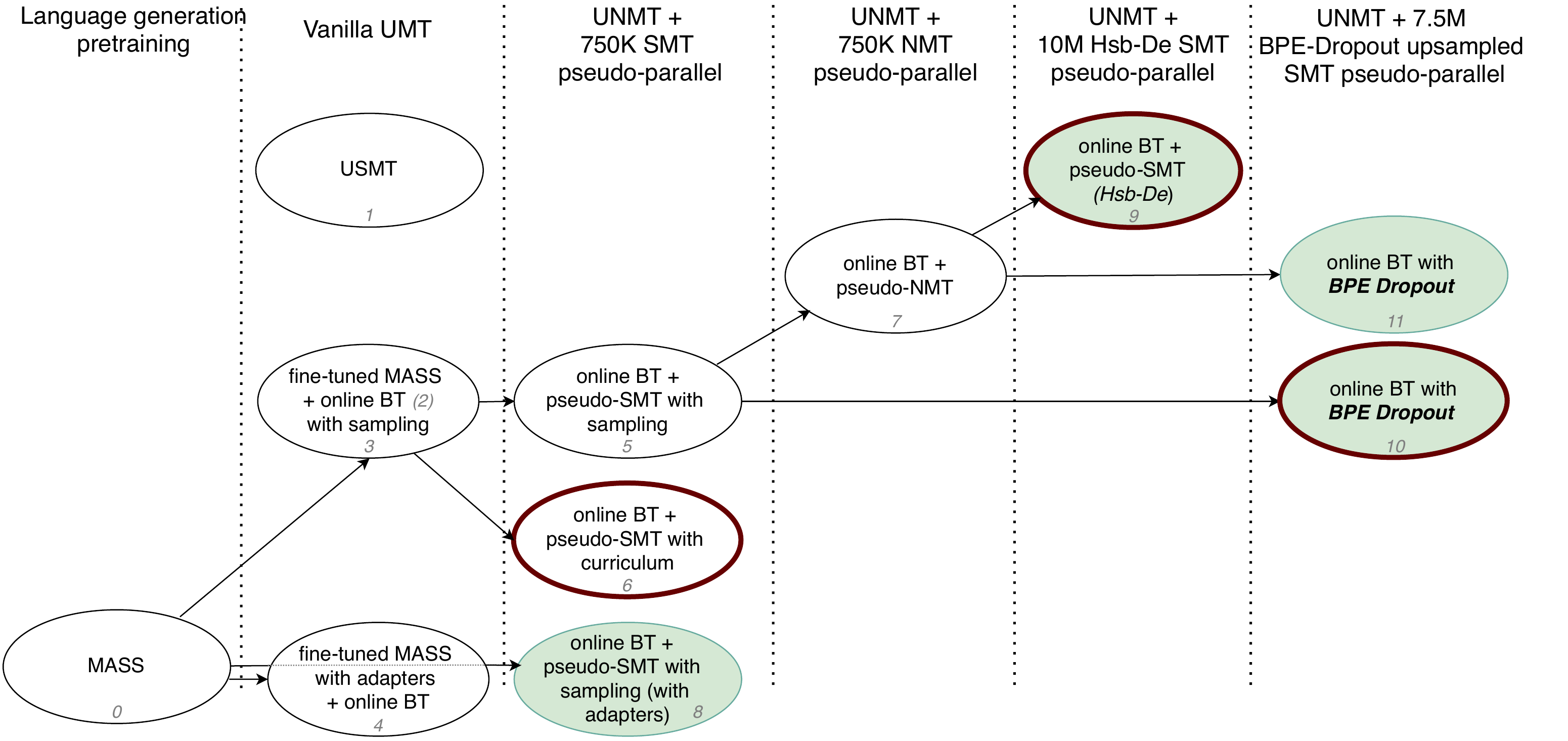}
	\caption{Illustration of our system. We denote with green the systems that were ensembled for the De$\rightarrow$Hsb direction and with maroon the systems that were ensembled for the Hsb$\rightarrow$De direction. Right arrows indicate transfer of weights. The numbers in gray correspond to the rows of Table \ref{table:dehsb}. 
	\texttt{Online BT} refers to the backtranslation of sentences with the actual model and updating it with the generated pseudo-parallel data. 
	\texttt{Pseudo-\textsc{smt}} refers to data obtained by backtranslating using the \textsc{usmt} baseline system while pseudo-\textsc{nmt} to our translations using system 5. The components of our approach are explained in Section \ref{sec:system}. }\label{fig:wmt}
\end{figure*}

Figure~\ref{fig:wmt} presents all the different components of our system and how they are connected to each other. 
We train both an unsupervised \textsc{smt} (\#1) and \textsc{nmt} (\#2) model. The \textsc{unmt} model is based on a pretrained \textsc{mass} model  (\#0), which is \textit{monolingual} (\texttt{De}). The model is later fine-tuned on both \texttt{Hsb} and \texttt{De}. We additionally explore fine-tuning only on \texttt{Hsb} using adapters. These models are used to initialize an \textsc{nmt} model (\#2, \#4) which is trained with online backtranslation. We additionally experiment with sampling (\#3) during backtranslation. The \textsc{usmt} model is used to backtranslate \texttt{Hsb} and \texttt{De} data. This synthetic bi-text is used to fine-tune the baseline \textsc{unmt} model (\#5). We use the synthetic bi-text also to fine-tune directly the adapter-augmented \textsc{mass} model, while employing online backtranslation and sampling (\#8).
We experiment with curriculum learning (\#6) to estimate the optimal way to feed the model this pseudo-parallel data. We also use our \textsc{unmt} model to generate backtranslations and fine-tune existing models (\#7). Further \textsc{usmt}-backtranslated data is used in \#9. Finally, some models are fine-tuned with monolingual data which is oversampled and segmented with BPE-Dropout (\#10, \#11). 
The details of these components are outlined in the following.  

\subsection{Unsupervised SMT}
\label{ssec: usmt}
First we describe the \textsc{usmt} system which we use to generate pseudo-parallel data to fine-tune our \textsc{nmt} system.
We use \emph{monoses} \cite{artetxe-etal-2018-unsupervised}, which builds unsupervised bilingual word embeddings (\textsc{bwe}s) and integrates them to Moses \cite{koehn2006open}, but apply some modifications to it.

As a first step, we build unsupervised \textsc{bwe}s with \emph{fastText} \cite{Bojanowski2017} and \emph{VecMap} \cite{Artetxe2018:vecmap} containing representations of $1$-, $2$- and $3$-grams.
Since the size of the available monolingual \texttt{Hsb} data is low, mapping monolingual embeddings to \textsc{bwe}s without any bilingual signal fails, i.e., we find no meaningful translations by manually investigating the most similar cross-lingual pairs of a few words.
Instead, we rely on identical words occurring in both  \texttt{De}  and \texttt{Hsb} corpora as the initial seed dictionary.
The \textsc{bwe}s are then converted to phrase-tables using cosine similarity of words and a language model is trained on the available monolingual data.
The shared task organizers released a validation set which we use to tune the parameters of the system with MERT, instead of running unsupervised tuning as described in \citet{artetxe-etal-2018-unsupervised}. Finally, we run 4 iterative refinement steps to further improve the system. Other than the above, all steps and parameters are unchanged.

We use this system in inference mode to backtranslate 
$7$M \texttt{De} and $750$K \texttt{Hsb} sentences. We refer to this pseudo-parallel dataset as \texttt{7.7M SMT pseudo-parallel}. 
We also backtranslate $10$M more \texttt{De} sentences. This dataset is later used to fine-tune one of our systems. We refer to it as \texttt{10M Hsb-De SMT pseudo-parallel}.

\subsection{MASS}
We initialize our \textsc{unmt} systems with an encoder-decoder Transformer \cite{vaswani2017attention}, which is pretrained using the \textsc{mass} \cite{song2019mass} objective. The model is pretrained by trying to reconstruct a sentence fragment given the remaining part of the sentence. The encoder takes a randomly masked fragment as input, while the decoder tries to predict the masked fragment. \textsc{mass} is inspired by \textsc{bert} \cite{devlin-etal-2019-bert}, but is more suitable for machine translation, as it pretrains the encoder-decoder and the attention mechanism, whereas \textsc{bert} is an encoder Transformer. In order to pretrain the model, instead of training  \textsc{mass} on both \texttt{De}  and \texttt{Hsb}, we initially train it on  \texttt{De}. After this, we fine-tune it on both \texttt{De}  and \texttt{Hsb}, following \textsc{re-lm} \cite{chronopoulou2020reusing}.
The intuition behind this is that, if we simultaneously train a cross-lingual model on unbalanced data, where $X$ is much larger than $Y$, the model starts to overfit the low-resource side $Y$ before being trained on all the high-resource language data ($X$). This results in poor translations. We refer to our pretrained model as \textsc{fine-tuned mass}.

\subsubsection{Vocabulary Extension for NMT}
\label{sssec:voctrick}
To fine-tune the pretrained  \texttt{De} \textsc{mass} model on \texttt{Hsb}, we need to overcome the following issue: the pretrained model uses BPE segmentation and vocabulary based only on \texttt{De}. To this end, we again follow \textsc{re-lm}.
We denote these BPE tokens as $\text{BPE}_{De}$ and the resulting vocabulary as $V_{De}$. We aim to fine-tune the monolingual \textsc{mass} model to \texttt{Hsb}. 
Splitting \texttt{Hsb} with $\text{BPE}_{De}$ would result in heavy segmentation of  \texttt{Hsb} words. To prevent this from happening, we learn BPEs on the joint  \texttt{De}  and \texttt{Hsb} corpus ($\text{BPE}_{joint}$).
We then use $\text{BPE}_{joint}$ tokens to split the \texttt{Hsb} data, resulting in a vocabulary $V_{Hsb}$. This method increases the number of shared tokens and enables cross-lingual transfer of the pretrained model. The final vocabulary is the union of the $V_{De}$ and $V_{Hsb}$ vocabularies. 
We extend the input and output embedding layer to account for the new vocabulary items. The new parameters are then learned during fine-tuning. 

\subsection{Adapters}
Besides initializing our \textsc{unmt} systems with \textsc{fine-tuned mass}, we also experiment with pretraining \textsc{mass} on  \texttt{De}  and fine-tuning \textit{only} on \texttt{Hsb}. During fine-tuning, we freeze the encoder and decoder Transformer layers and add adapters \cite{Houlsby2019ParameterEfficientTL} to each of the Transformer layers. Adapters can prevent catastrophic forgetting \cite{goodfellow2013empirical} and show promising results in various tasks \cite{bapna-firat-2019-simple,artetxe2019cross}. We fine-tune only the output layer, the embeddings and the decoder's attention to the encoder as well as the lightweight adapter layers. 

We investigate adapters as fine-tuning in this way is considerably more computationally efficient. We also experimented with freezing the decoder's attention to the encoder as well as adding an adapter on top of it, but these architecture designs are worse in terms of 
perplexity during \textsc{mass} fine-tuning as well as 
BLEU scores during \textsc{unmt}. 

We use the fine-tuned model to initialize an encoder-decoder Transformer, augmented with adapters. The adapter-augmented model is then trained in an unsupervised way, using online backtranslation. All layers are trainable during unsupervised \textsc{nmt} training.
 We refer to this model as \textsc{fine-tuned mass + adapters}.
 
\subsection{Unsupervised NMT (online backtranslation)}
We initialize our \textsc{unmt} models with \textsc{fine-tuned mass}. Following \newcite{song2019mass}, we train the systems in an unsupervised manner, using online
backtranslation \cite{sennrich2015improving} of the monolingual \texttt{Hsb} and  \texttt{De}  data, that were also used for pretraining. 
As proposed in \newcite{song2019mass}, we do not use denoising auto-encoding \cite{vincent2008extracting}.
We use online
backtranslation to generate pseudo bilingual data for training. 
We refer to the resulting model as \textsc{unmt baseline}.

\subsection{Sampling}
\label{ssec:sampling}
\enote{DS}{Moving this after UNMT}

We experiment with sampling instead of greedy decoding during 
online
backtranslation. \newcite{edunov2018understanding} show that sampling is beneficial for backtranslation compared to greedy decoding or beam search for systems trained on larger amounts of parallel data. Although we do not use any parallel data, we assumed that our initial \textsc{unmt} baseline is of reasonable quality and that sampling would be beneficial. However, in order to provide a balance, we randomly use either greedy decoding or sampling during training. The frequency with which sampling is used is a hyperparameter which we set to $0.5$. Sampling temperature 
is set to $0.95$. 

\enote{DS}{Commenting the previous paragraph. Not sure if we should discuss this, I now see that the optimal values are not the ones we chose, but we went with these because: we already had a model trained, the differences were not that large, training on 1 gpu made the results worse, so not very reliable.}

\subsection{Curriculum learning}
\label{ssec:cl}

\enote{DS}{This is probably too long, will have to cut it.}

Considering the high improvements achieved by including \textsc{smt} backtranslated data, we conduct experiments to determine  a more meaningful way to feed the data to the model 
using curriculum learning \cite{kocmi-bojar2017curriculum,platanios2019competence,zhang2019curriculum}.
We learn the curriculum using Bayesian Optimization (BO) for which we use an open source implementation\footnote{\url{https://ax.dev/}}. Similar work has been proposed for transfer learning \cite{ruder-plank2017learning} and \textsc{nmt} \cite{wang2020learning-multi}. As we already have a reasonably trained \textsc{nmt} model, we use it to compute instance-level features for learning the curriculum. Each sentence pair from the \textsc{smt} backtranslated data is represented with two features: the model scores for this pair in the \textit{original} (backtranslation $\rightarrow$ monolingual sentence) and \textit{reverse} direction (monolingual $\rightarrow$ backtranslation).

The weights that determine the importance of these features are learned separately for  \texttt{De}$\rightarrow$\texttt{Hsb} and \texttt{Hsb}$\rightarrow$\texttt{De}, so that we have $4$ features in total.
BO runs for $30$ trials. The feature weights are constrained in the range [$-1$, $1$]. Each trial runs $5.4$K \textsc{nmt} updates. The curriculum optimizes the sum of \texttt{Hsb}$\rightarrow$\texttt{De} and \texttt{De}$\rightarrow$\texttt{Hsb} validation perplexity. 
For the optimization trials, we only use the \textsc{smt} backtranslated data as pseudo-parallel data and do not use 
online backtranslation. 
Finally, based on the feature weights and the features for each sentence, we sort the pseudo-parallel data and fine-tune the \textsc{unmt baseline} with \textsc{smt} backtranslations and 
online
backtranslation. It would be interesting to study if a similar approach can be used to estimate a more optimal loading of monolingual data during \textsc{mass} pretraining and \textsc{unmt}. 

\subsection{Offline Iterative Backtranslation}
We also experiment with creating synthetic training data using offline backtranslation with one of our \textsc{unmt} systems (\#5 in Table \ref{table:dehsb}). We translate $750$K \texttt{De} sentences to  \texttt{Hsb} and $750$K  \texttt{Hsb} sentences to De. The resulting pseudo-parallel system is denoted as \texttt{750K NMT pseudo-parallel} corpus and is used to fine-tune the same system. 

\subsection{BPE-Dropout}
\label{ssec:bpedrop}
BPE segmentation is useful in machine translation, as it efficiently addresses the open vocabulary problem. This approach keeps the most frequent words intact and splits the rare ones into multiple tokens. It builds a vocabulary of subwords and a merge table, specifying which subwords have to be merged and the priority of the merges.
BPE segmentation always splits a word  deterministically. Introducing stochasticity to the algorithm \cite{bpedropout}, by simply removing a merge from the merges with a pre-defined probability $p$, results in significant BLEU improvements for various languages in low- and medium-resource datasets. 

We use BPE-Dropout in the following way: we oversample
the \texttt{Hsb} monolingual data by a factor of $10$ and apply BPE-Dropout.
In that way, we get different segmentations of the same sentences and feed this
data to the model. We also oversample the \texttt{750K SMT pseudo-parallel} corpus 
in the same manner, but only apply BPE-Dropout to the \texttt{Hsb} side. These monolingual and pseudo-parallel 
oversampled datasets are used to fine-tune our models. 
These systems perform better than our other single systems. 

\enote{DS}{Alternative place to describe system}

\subsection{Ensembling}
For the final models, we perform ensemble decoding with the best training models obtained in our experiments.
We evaluate several combinations of model ensembles. 
Based on BLEU scores on the test set provided during development, we decide on two separate ensembles for \texttt{De}$\rightarrow$\texttt{Hsb} and \texttt{Hsb}$\rightarrow$\texttt{De} for the final submission. 
\enote{DS}{Moving this to Results, probably it would be easier to understand because we mention quotation issue in Experiments}

\begin{table*}[ht]
\centering
\small
\begin{tabular}{clrr}
\toprule
 \textbf{\#}& \textbf{Methods} &\textbf{ De$\rightarrow$Hsb}  &\textbf{ Hsb$\rightarrow$De} \\ 
 \midrule
 0 & MASS & 5.6 & 7.0 \\
 1 & USMT &  19.3 & 21.4\\ 
 2 & \raisebox{.5pt}{\textcircled{\raisebox{-.9pt} {0}}} UNMT baseline (fine-tuned MASS) &  24.4  &  27.1  \\ 
 3 & \raisebox{.5pt}{\textcircled{\raisebox{-.9pt} {2}}} UNMT baseline  + sampling       &  25.4  &  27.4  \\ 
 4 & \raisebox{.5pt}{\textcircled{\raisebox{-.9pt} {0}}} UNMT baseline (fine-tuned MASS with adapters) & 18.8 & 21.7 \\  \midrule
  5 & \raisebox{.5pt}{\textcircled{\raisebox{-.9pt} {3}}} +  online BT + pseudo-SMT + sampling & 29.9 & 31.9  \\
 6 & \raisebox{.5pt}{\textcircled{\raisebox{-.9pt} {3}}}
+ online BT + pseudo-SMT + curriculum & \underline{30.0} & 32.5 \\
\enote{DS}{}
 \phantom{*}6* & \raisebox{.5pt}{\textcircled{\raisebox{-.9pt} {3}}}
+ online BT + pseudo-SMT + curriculum + sampling & 30.2 & 32.8 \\
 7 & \raisebox{.5pt}{\textcircled{\raisebox{-.9pt} {5}}} + online BT + pseudo-NMT & 29.8 & \underline{33.2} \\
 8  & \raisebox{.5pt}{\textcircled{\raisebox{-.9pt} {0}}} + online BT + pseudo-SMT + sampling (with adapters) & 29.0 & 32.3 \\
 9 & \raisebox{.5pt}{\textcircled{\raisebox{-.9pt} {7}}} + online BT + pseudo-SMT (Hsb-De)  & \underline{30.0} & 32.7 \\
 \midrule
\multicolumn{3}{c}{Data oversampling with BPE-Dropout} \\  \midrule 
 10 & \raisebox{.5pt}{\textcircled{\raisebox{-.9pt} {5}}} + BPE-Dropout & 30.7 & 33.4 \\
 11 &\raisebox{.5pt}{\textcircled{\raisebox{-.9pt} {7}}} + BPE-Dropout & \underline{31.8} &  \underline{34.0} \\
 \midrule 

 12 & Model Ensemble (8, 9, 10, 11) & \textbf{32.4} & \textbf{35.2} \\
 13 & Model Ensemble (6, 9, 11) & 31.9 & 34.8 \\

\bottomrule

\end{tabular}
\caption{BLEU scores of UMT for De-Hsb and Hsb-De systems. The systems with the underlined results were ensembled and used in our primary submissions. \#12 is our primary system submitted to the organizers in the De$\rightarrow$Hsb direction, while \#13 is our primary system submitted in the Hsb$\rightarrow$De direction. 6* was trained after the shared task and is not used for the final submission.}

\label{table:dehsb}
\end{table*}

\section{Experiments}

\subsection{Data Pre-processing}
In line with the rules of the WMT $2020$ unsupervised shared task\footnote{\url{http://www.statmt.org/wmt20/unsup_and_very_low_res/}}, we used $327$M sentences from WMT monolingual News Crawl\footnote{\url{http://data.statmt.org/news-crawl/de/}} dataset for German, collected over the period of $2007$ to $2019$. We also used the Upper Sorbian side of the provided parallel data as well as all of the  monolingual data, a total amount of $756$K sentences, provided by the organizers. We used the provided parallel data for validation/testing  ($2$K/$2$K sentences). We normalized punctuation, tokenized and true-cased the data using standard scripts from the Moses toolkit \cite{koehn2006open}. We note that we tokenized  \texttt{Hsb} data using Czech as the language of tokenization, since these two languages are very closely related and there are no tokenization rules for \texttt{Hsb} in Moses.

We used BPE \cite{sennrich2015neural} segmentation for our neural system. Specifically, we learned $32$K codes and computed the vocabulary using the \texttt{De} data. We then also learned the same amount of BPEs on the joint corpus (\texttt{De}, \texttt{Hsb}) and computed the joint vocabulary. We extended the initial vocabulary, adding to it unseen items. We used this augmented vocabulary to fine-tune the \textsc{mass} model and run all the \textsc{unmt} training experiments. 

\subsection{Data Post-processing}
We fixed the quotes to be the same as in the source sentences (German-style). We also applied a recaser using Moses \cite{koehn2006open} to convert the translations to 
mixed case.

\subsection{Training}

\noindent\textbf{Unsupervised \textsc{smt}.}
As mentioned before, we used \textit{fastText} \cite{Bojanowski2017} to build $300$ dimensional embeddings on the available monolingual data.
We build \textsc{bwe}s with \textit{VecMap} \cite{Artetxe2018:vecmap} using identical words as the seed dictionary and restricting the vocabulary to the most frequent $200$K, $400$K and $400$K $1$-, $2$- and $3$-grams respectively.
We used \emph{monoses} \cite{artetxe-etal-2018-unsupervised} as the \textsc{usmt} pipeline but used the available validation data for parameter tuning and ran $4$ iterative refinement steps.

\noindent\textbf{\textsc{mass}.}
We use a Transformer, which consists of $6$-layer encoder and $6$-layer decoder with $1024$ embedding/hidden size, $4096$ feed-forward network size and $8$ attention heads. We pretrain \textsc{mass} on \texttt{De} monolingual data, using Adam \cite{kingma2014adam} optimizer with inverse square root learning rate scheduling and a learning rate of $10^{-4}$. We used a per-GPU batch size of $32$. We trained the model for approximately $2$ weeks on $8$ NVIDIA GTX $1080$ Ti $11$ GB GPUs. The rest of the hyperparameters follows the original \textsc{mass} paper. We fine-tune \textsc{mass} on both  \texttt{De}  and \texttt{Hsb} using the same setup, but on $4$ GPUs of the same type. Fine-tuning was performed for $2$ days.

\noindent\textbf{Unsupervised \textsc{nmt}.}
For unsupervised \textsc{nmt}, we further train the fine-tuned MASS using online backtranslation. We use 4 GPUs to train each one of our UNMT models. 
We report 
BLEU using SacreBLEU \cite{post-2018-call}\footnote{BLEU+case.mixed+numrefs.1+smooth.exp+tok.13a+ver\-sion.1.4.13} on the provided test set.

\noindent\textbf{Unsupervised \textsc{nmt} + Pseudo-parallel \textsc{mt}.} We train our \textsc{unmt} systems using a pseudo-parallel supervised translation loss, in addition to the online backtranslation objective. We found out that augmenting \textsc{unmt} systems with pseudo-parallel data obtained by \textsc{usmt} leads to major improvements in translation quality, as previous work has showed \cite{artetxe-etal-2018-unsupervised,stojanovski-etal-2019-lmu}.

\section{Results}
The results of our systems on the test set provided during development are presented in Table \ref{table:dehsb}. 
Our \textsc{usmt} model (\#1) performs competitively, but is largely outperformed by the \textsc{unmt} baseline (\#2). 
These results are interesting considering that both systems are trained using small amounts of monolingual \texttt{Hsb} data. 
We believe that the performance of the \textsc{unmt} model is largely due to the \textsc{mass} fine-tuning scheme which allowed us to obtain a strong pretrained model for both languages.
We also observe (\#3) that mixing greedy decoding and sampling during backtranslation is beneficial compared to always using greedy decoding (\#2), especially for  \texttt{De} $\rightarrow$\texttt{Hsb} which improved by $1.0$ BLEU. 
However,
it is likely that sampling is useful only if the model is of reasonable quality. We note  that the adapter-augmented model (\#4) is worse than the \textsc{unmt} baseline.

After these initial experiments, we use the \textsc{usmt} model (\#1) to backtranslate all \texttt{Hsb} monolingual data and $7$M  \texttt{De} sentences. This pseudo-parallel data is leveraged to fine-tune our \textsc{unmt} models alongside online backtranslation.
This approach, denoted as model \#5, improves the \textsc{unmt} baseline (\#3) by more than $5.5$ BLEU for \texttt{De}$\rightarrow$\texttt{Hsb} and $4.5$ BLEU for \texttt{Hsb}$\rightarrow$\texttt{De}. 
The curriculum learning approach (\#6) yields a small improvement of $0.6$ BLEU for \texttt{Hsb}$\rightarrow$\texttt{De}.
Unfortunately, the curriculum learning model ran without the use of sampling. 
We later train the model with sampling (\#6*) and obtain slight improvements in both directions.

Using \textsc{nmt} backtranslations in an offline manner (\#7) provides for a large improvement in the \texttt{Hsb}$\rightarrow$\texttt{De} direction, obtaining $33.2$ BLEU. 
Further training our high scoring model \#7 on \textsc{usmt} backtranslations, depicted as model \#9, degrades performance on \texttt{Hsb}$\rightarrow$\texttt{De}. This might indicate that \textsc{usmt} backtranslations alone are not very important for high performance, but simply adding any kind of pseudo-parallel data during training. 

The adapter-augmented model with \textsc{usmt} backtranslations (\#8) manages to close the gap to the baseline model. Comparing \#5 and \#8, we can see that the model with adapters is worse by $0.9$ BLEU on \texttt{De}$\rightarrow$\texttt{Hsb}, but 
better by $0.4$ on \texttt{Hsb}$\rightarrow$\texttt{De}. 
Due to time constraints, we train \#4 and \#8 in parallel and \#8 is not fine-tuned from \#4. 
Overall, adapters are a promising research direction as they lead to faster \textsc{mass} fine-tuning and comparable performance.  

We observe considerable improvements using BPE-Dropout. As noted before, we oversample the parallel and \texttt{Hsb} monolingual data and apply BPE-Dropout only on \texttt{Hsb}. We use this data to fine-tune some of our already trained models, specifically \#5 and \#7 which results in models \#10 and \#11, respectively. This approach improves the  \texttt{Hsb}$\rightarrow$\texttt{De} direction by up to $1.5$ BLEU and up to $1.0$ BLEU for \texttt{De}$\rightarrow$\texttt{Hsb}. 
System \#11 proved to be our best single system in both translation directions.
We hypothesize that using BPE-Dropout while simultaneously oversampling the data provides for a data augmentation effect.
In future work, it would be interesting to decouple these two steps and measure their effect separately.

Ensembling further boosts performance. 
Ensemble \#12 is used for \texttt{De}$\rightarrow$\texttt{Hsb} and \#13 for \texttt{Hsb}$\rightarrow$\texttt{De}.
We note that while computing ensemble BLEU scores during development, we did not fix the issue with German-style quotes. This resulted in ensemble \#13 obtaining better scores on \texttt{Hsb}$\rightarrow$\texttt{De}.
We later fix the quotes issue and find out that ensemble \#12 is better on both translation directions and is the best system overall. 

\section{Conclusion}
In this paper, we present the LMU Munich system for the WMT $2020$ unsupervised shared task for translation between German and Upper Sorbian. Our system is a combination of an \textsc{smt} and an \textsc{nmt} model trained in an unsupervised way. The \textsc{unmt} model is trained by fine-tuning a \textsc{mass} model, according to the recently proposed \textsc{re-lm} approach.
The experiments show that the \textsc{mass} fine-tuning technique is efficient even if little monolingual data is available for one language and results in a strong \textsc{unmt} model. We also show that using pseudo-parallel data from \textsc{usmt} and \textsc{unmt} backtranslations improves performance considerably. Furthermore, we show that oversampling the low-resource Upper Sorbian and applying BPE-Dropout, which can effectively be seen as data augmentation, results in further improvements. 
Adapters in \textsc{mass} fine-tuning provided for a balance between performance and computational efficiency.
Finally, smaller but noticeable gains are obtained from using curriculum learning and sampling during decoding in backtranslation.

\section*{Acknowledgments}

This work was supported by the European Research Council (ERC) under the
European Union’s Horizon $2020$ research and innovation programme (grant
agreement No.~$640550$) and by the German Research Foundation (DFG; grant FR
$2829$/$4$-$1$). We would like to thank Jind\v{r}ich Libovick\'{y} for fruitful discussions regarding the use of BPE-Dropout as a data augmentation technique.

\bibliographystyle{acl_natbib}
\bibliography{emnlp2020}

\end{document}